\documentclass{article}
\usepackage[utf8]{inputenc}
\usepackage[margin = 1in]{geometry}

\usepackage{amsmath, amssymb, amsthm, amsfonts, mathtools, multirow, mathrsfs, tabu, booktabs, url, subcaption, graphicx,xcolor, tabu,  authblk, hyperref, blindtext}

\usepackage[english]{babel}
\usepackage[shortlabels]{enumitem}
 \usepackage[ruled,vlined, linesnumbered]{algorithm2e}
\usepackage[capbesideposition=outside,capbesidesep=quad]{floatrow}
\usepackage[framemethod=tikz]{mdframed}
\usepackage{bbm}

\usepackage{algorithmicx}
\usepackage{afterpage}  % for '\afterpage' command

\usepackage[backend=biber,style=ieee,sorting=none,natbib=true]{biblatex} 
\usepackage[autostyle=true, threshold=2]{csquotes} 

\title{\textbf{PalmProbNet: A Probabilistic Approach to Understanding Palm Distributions in Ecuadorian Tropical Forest via Transfer Learning}}
\author[1]{Kangning Cui\footnote{Corresponding Author: kangnicui2-c@my.cityu.edu.hk}}
\author[2]{Zishan Shao}
\author[2]{Gregory Larsen}
\author[2]{Victor Pauca}
\author[2]{Sarra Alqahtani}
\author[2]{David Segurado}
\author[2]{Jo\~{a}o Pinheiro}
\author[2]{Manqi Wang}
\author[3]{David Lutz}
\author[2]{Robert Plemmons}
\author[2]{Miles Silman}
\affil[1]{ City University of Hong Kong, Kowloon, Hong Kong}
\affil[2]{ Wake Forest University, Winston-Salem, NC, USA}
\affil[3]{ Dartmouth College, Hanover, NH, USA}
\date{}                     %% if you don't need date to appear

\begin{document}
\topmargin=0mm
\maketitle
\begin{abstract}

Palms play an outsized role in tropical forests and are important resources for humans and wildlife. A central question in tropical ecosystems is understanding palm distribution and abundance. However, accurately identifying and localizing palms in geospatial imagery presents significant challenges due to dense vegetation, overlapping canopies, and variable lighting conditions in mixed-forest landscapes. Addressing this, we introduce PalmProbNet, a probabilistic approach utilizing transfer learning to analyze high-resolution UAV-derived orthomosaic imagery, enabling the detection of palm trees within the dense canopy of the Ecuadorian Rainforest. This approach represents a substantial advancement in automated palm detection, effectively pinpointing palm presence and locality in mixed tropical rainforests. Our process begins by generating an orthomosaic image from UAV images, from which we extract and label palm and non-palm image patches in two distinct sizes. These patches are then used to train models with an identical architecture, consisting of an unaltered pre-trained ResNet-18 and a Multilayer Perceptron (MLP) with specifically trained parameters. Subsequently, PalmProbNet employs a sliding window technique on the landscape orthomosaic, using both small and large window sizes to generate a probability heatmap. This heatmap effectively visualizes the distribution of palms, showcasing the scalability and adaptability of our approach in various forest densities. Despite the challenging terrain, our method demonstrated remarkable performance, achieving an accuracy of 97.32\% and a Cohen's $\kappa$ of 94.59\% in testing.

\end{abstract}


\section{Introduction}

Detecting palm trees in tropical forests is important in both biological and computational domains. Biologically, palms serve as important ecological indicators, providing insights into biodiversity, soil quality, and overall health of the forest ecosystem, support human livelihoods in indigenous and rural communities, and are keystone resources for tropical wildlife~\cite{kipli2023deep, chong2017review}. Computationally, particularly within image processing and machine learning, palm detection is challenging due to issues such as noise and artifacts in image orthomosaics, variable illumination within forests, lack of labeled data, and  data imbalances \cite{kipli2023deep, chong2017review, khan2021oil}. Addressing these challenges is essential not only for ecological studies that rely on identification and localization of forest resources but also for advancing computer vision techniques for remote sensing in general.

In response to the outlined challenges, this study introduces PalmProbNet, a deep learning approach tailored for palm tree detection in Ecuador's tropical forests, capitalizing on UAV-derived orthomosaic imagery. We employ a strategy of training two models, both with a consistent architecture, on image patches of varying sizes, resulting in probability maps that depict palm distributions. Experimental results highlight the robustness and efficacy of PalmProbNet in classifying palm trees. \underline{Our contributions are threefold}: a data labeling process that categorizes two sets of image patches containing palm and non-palm characteristics in different sizes, feature extraction via transfer learning integrated with deep learning-based classification, and the application to the full UAV-derived orthomosaic imagery. The results emphasize the promise of integrating UAV technology with deep learning  for efficient palm tree detection within dense forest canopies. 

The subsequent sections of the paper are structured as follows: Section 2 provides a comprehensive review of related work in palm detection and deep learning. Section 3 details our dataset, including data collection, preprocessing, and labeling. Section 4 introduces PalmProbNet, our approach covering feature extraction, classification, and application to the landscape orthomosaic image. Section 5 showcases the experimental setup and its findings, succeeded by a comprehensive discussion on these outcomes. Finally, Section 6 concludes the paper and outlines directions for future research.

\section{Related Work}

\subsection{Object Detection}

Object detection, a pivotal task in computer vision, identifies and classifies objects within images. Broadly, the methodologies can be categorized into traditional and deep learning-based approaches \cite{zhao2019object, devi2021review, zou2023object, wang2019automatic, cui2022semi}. Traditional approaches follow a structured pipeline: selecting informative regions, extracting hand-crafted features, and classifying these features. Specifically, regions of interest (RoIs) within the image, which contains the targeted objects, are identified using methods like sliding window or selective search \cite{wang2019automatic, khuzaimah2022application}. The selected RoIs are then subjected to feature extraction approaches to describe specific features within the region, capturing object appearance, shape, and texture information~\cite{wang2019automatic, zou2023object, malek2014efficient}. The features are then fed into classifiers to categorize the RoIs and finalize the object detection~\cite{wang2019automatic, devi2021review}. Traditional classifiers tend to be computationally efficient due to their fewer tunable parameters, in contrast to their deep learning counterparts \cite{wang2019automatic, zou2023object}. 

Deep learning-based techniques, on the other hand, offer an end-to-end training and prediction process by integrating feature extraction and classification into a unified framework~\cite{kipli2023deep, zhao2019object, camalan2022change, marin2022aerial}. Notable deep learning algorithms include You Only Look Once (YOLO) and Region-Based Convolutional Neural Networks (R-CNNs)~\cite{zou2023object}. YOLO adopts a single-shot architecture that enables it to sufficiently perform object detection in real-time, while R-CNNs leverage a two-step process that first identifies regions of interest and then classifies those regions, providing a more accurate but computationally intensive approach~\cite{zhao2019object, zou2023object, marin2022aerial, liu2021automatic}. Despite their capabilities, both traditional and deep learning methods encounter challenges in real-world scenarios, especially in detecting palm trees. Factors like dense canopies and diverse species in forests introduce complexities such as overlapping objects and intricate backgrounds \cite{osco2021review, diez2021deep}.

\begin{figure}[tb]
    \centering
    \begin{subfigure}{0.6\linewidth}
        \centering
        \includegraphics[width=\textwidth]{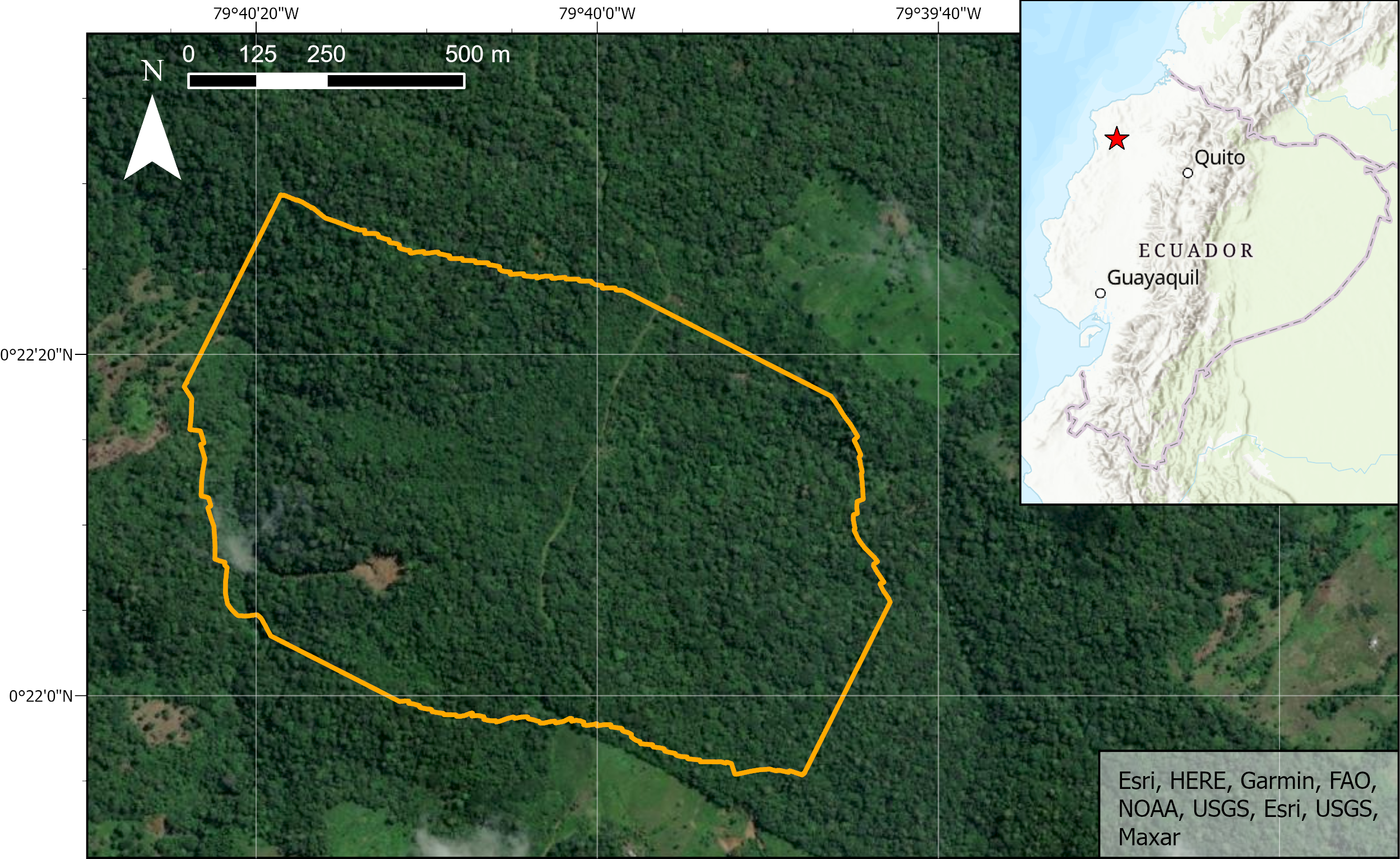}
        \caption{Map of the Study Area}
    \end{subfigure}
    
    \begin{subfigure}{0.3\linewidth}
        \centering
        \includegraphics[width=\textwidth]{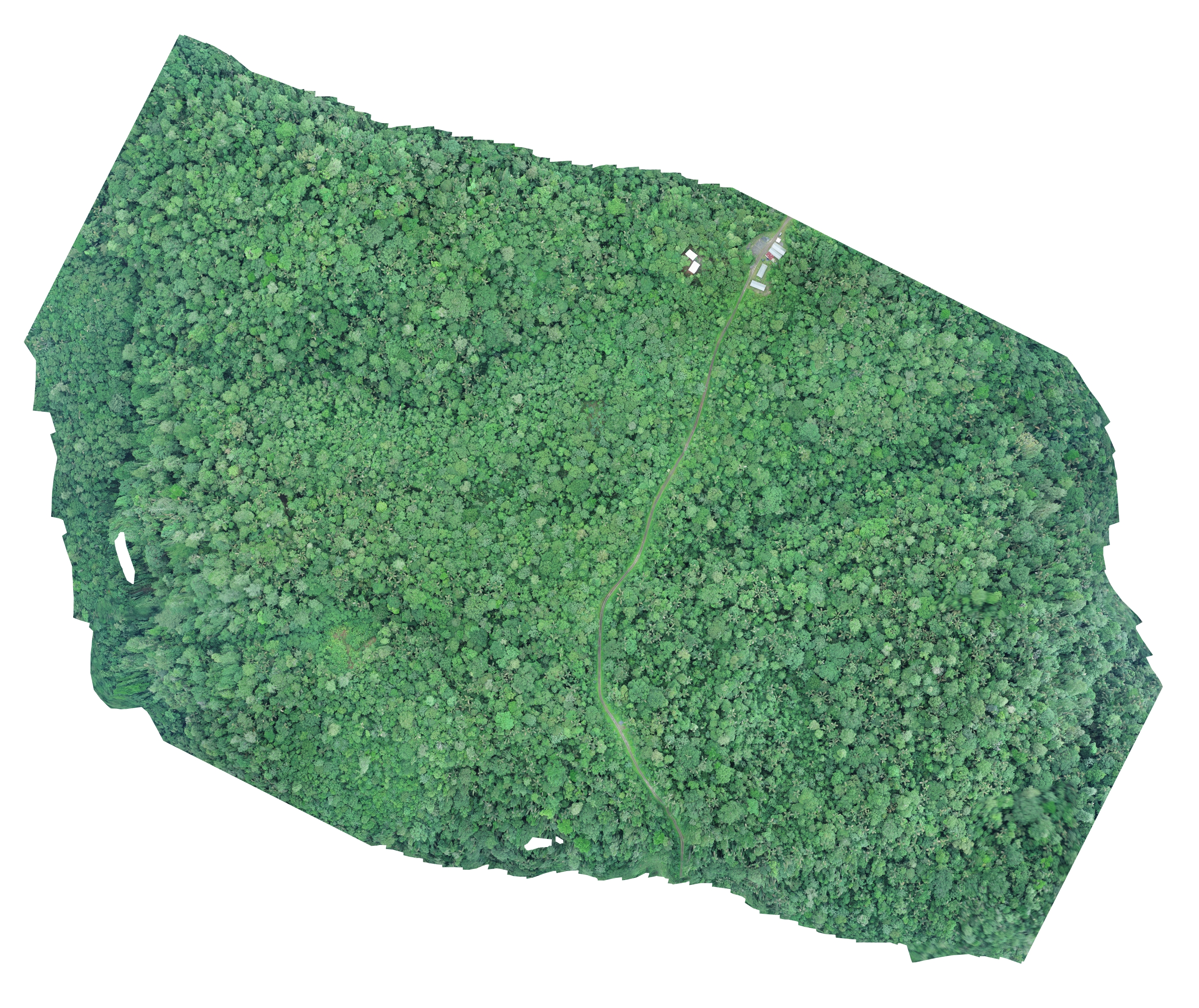}
        \caption{The Original Orthomosaic}
    \end{subfigure}
    \begin{subfigure}{0.3\linewidth}
        \centering
        \includegraphics[width=\textwidth]{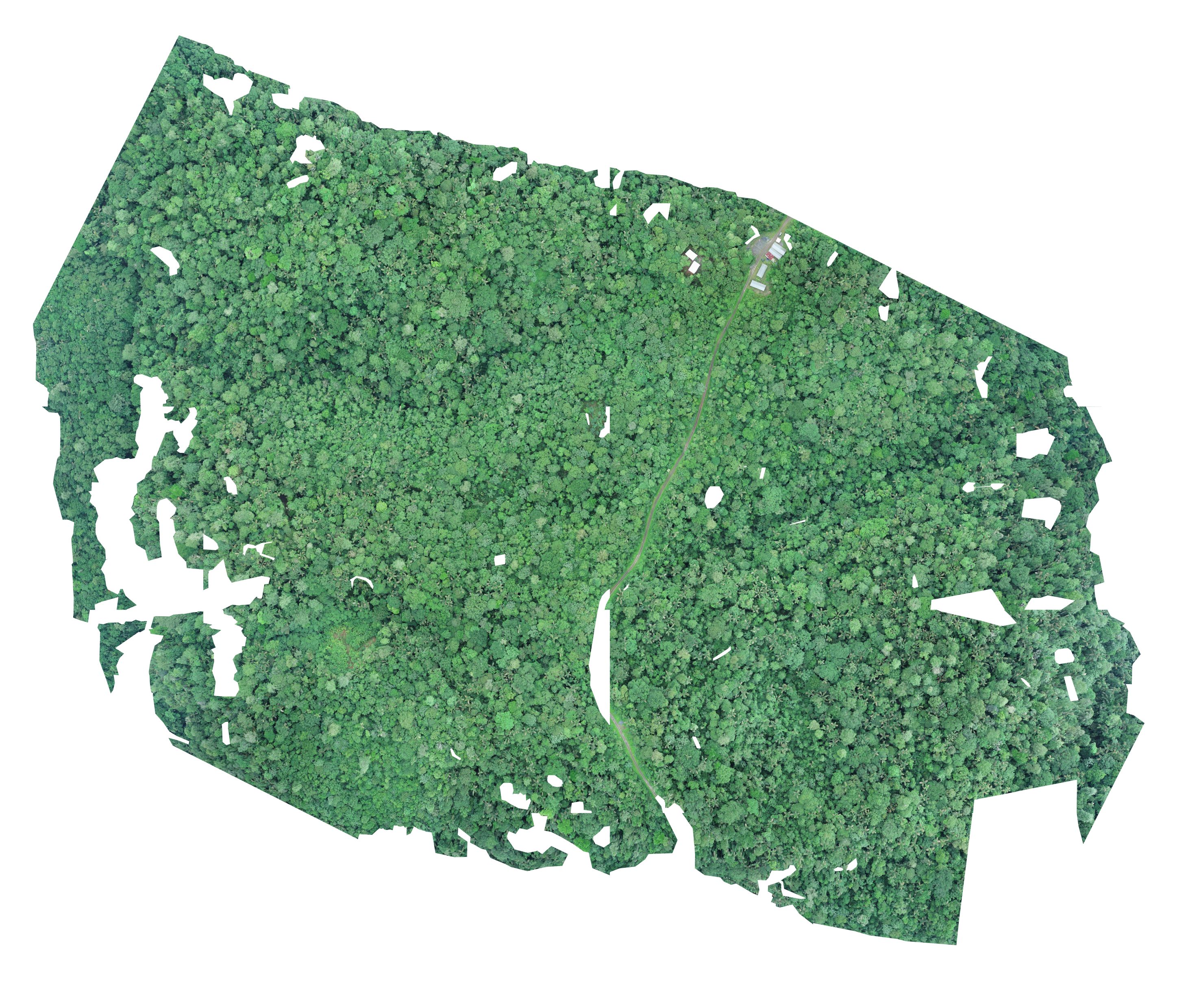}
        \caption{The Cleaned Orthomosaic}
    \end{subfigure}
    \caption{Visualizations of the Orthomosaic Image Before and After Cleaning}
    \label{fig:orthomosaic}
\end{figure}

\subsection{Palm Detection}

\begin{figure}[tb]
  \centering
  \begin{subfigure}{0.2\textwidth}
    \includegraphics[width=\textwidth]{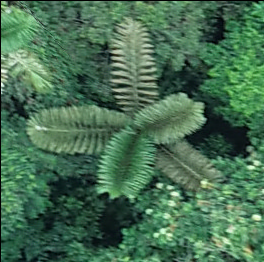}
    \caption{Isolated Palm}
  \end{subfigure}
  \begin{subfigure}{0.2\textwidth}
    \includegraphics[width=\textwidth]{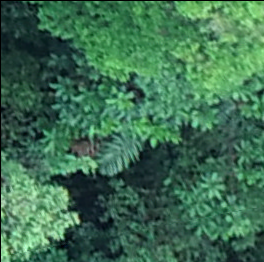}
    \caption{Hidden Palm}
  \end{subfigure}
  \begin{subfigure}{0.2\textwidth}
    \includegraphics[width=\textwidth]{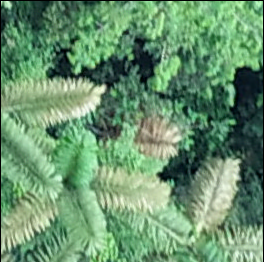}
    \caption{Overlapped Palms}
  \end{subfigure}
  \caption{Palms as Captured in the Orthomosaic}
  \label{fig:palm_images}
\end{figure}

Modern remote sensing technologies have significantly advanced palm tree detection by increasing the abundance, availability and resolutions of both satellite and aerial imagery. Remote sensing often leverages specialized sensors to capture distinct spectral and structural characteristics~\cite{kipli2023deep, osco2021review}. For example, thermal sensors monitor surface temperature and can aid in irrigation design for palm plantations, multispectral sensors produce NDVI images which can help distinguish the presence, types, and organismal qualities of vegetation, hyperspectral sensors yield detailed profiles of spectral reflectance across large segments of the electromagnetic spectrum, and LiDAR sensors provide topographic and structural data~\cite{khuzaimah2022application, khan2021oil}. These sensors are deployed on spaceborne platforms (satellites), airborne platforms (drones or occupied aircrafts) or ground-based platforms, with trade-offs of cost, scalability, and observation scale depending on the platform and its deployment \cite{khuzaimah2022application, kipli2023deep}. In this context, unoccupied aerial vehicles (UAVs) provide a distinctly scalable, cost-effective opportunity for canopy monitoring with ultra-high image resolutions, achievable due to their typically high-resolution sensors and ability to operate safely at very low altitudes \cite{khan2021oil, liu2021automatic}. UAVs enable efficient monitoring of vast extents of terrain, catering especially to needs of small-scale farmers monitoring diverse palm species \cite{khuzaimah2022application, wang2019automatic}. The integration of automated palm detection techniques, whether conventional or deep learning-based, further accentuates the potential of UAV applications for mapping and monitoring vegetation land use and land cover.

UAVs and machine learning approaches rapidly enhance vegetation analysis for agricultural and forestry management, with methods like sliding window, histogram of oriented gradients, and SVM effectively detecting palm trees in UAV imagery from Malaysia and Saudi Arabia~\cite{khan2021oil, wang2019automatic, malek2014efficient, wang2019automatic, malek2014efficient}. However, while effective, machine learning classifiers depend on feature quality. Introducing a new dataset necessitates fresh feature extraction, which complicates the training phase~\cite{fassnacht2016review}. In contrast, deep learning approaches, particularly CNNs, inherently and adaptively extract relevant features during training, remarkably simplifying the process. CNNs have been applied to canopy monitoring with UAVs, effectively addressing complex object detection challenges~\cite{osco2021review, marin2022aerial, diez2021deep, junos2021automatic}. R-CNNs can effectively identify Mauritia flexuosa palms in Amazonian forests and oil palm trees in Malaysian plantations, demonstrating CNNs' adaptability in both forest and agricultural settings~\cite{marin2022aerial, liu2021automatic}. Moreover, other architectures like YOLO have been successfully utilized to detect loose fruits of oil palms from UAV images, contributing a major advance to precision agriculture~\cite{junos2021automatic}.

Identifying palm trees from high-resolution aerial or satellite imagery presents several challenges. Conventional classifiers may not capture hierarchical feature representations adequately, whereas deep-learning methods demand extensive training datasets~\cite{zou2023object, khuzaimah2022application}. Integrating CNN-based methods with conventional classifiers can harness the strengths of both, offering a more robust solution for palm tree detection~\cite{zou2023object, devi2021review}. This has been demonstrated, using CNNs to detect oil palm trees in satellite images using sliding window and post-processing techniques~\cite{li2016deep}. 

\section{Dataset}

\subsection{Raw and Orthomosaic Data}

The raw data used in this study pertains to a natural reserve in the Northwest region of Ecuador, known as the Ecuadorian Choco forest ($00^\circ23'28''\,\text{N},\,79^\circ41'05''\,\text{W}$), see Figure \ref{fig:orthomosaic}. The reserve is a high diversity humid tropical forest at ~500m elevation, receiving $\sim$ 3000 mm precipitation per year. Precipitation is seasonal, with a 5-month minimum from July-November accompanied by persistent fog (see~\cite{lueder2022functional} for detailed description). The forests contain 16 species of palms that can have exposed crown~\cite{browne2016diversity, lueder2022functional}.

The data was collected in collaboration with the Fundaci\'{o}n para la Conservaci\'{o}n de los Andes Tropicales (FCAT). The UAV imagery was collected in two campaigns.  The first in June 2022 covered 95 ha using a DJI Phantom 4 RTK drone with a 1'' CMOS sensor and controlled using UgCS mission planning software (CITS).  Missions were flown at 90 $m/s$ with 70\% sidelap and 80\% endlap. A total of 387 photos were taken and an orthomosaic and 3D point cloud were generated using Structure from Motion (SfM) in Agisoft Metashape 2.0 (Agisoft 2022). Flight plans, raw imagery, and orthomosaics are available upon request.

An orthomosaic is a large photogrammetrically orthorectified image product created from an image collection of a particular geographical region. Orthomosaics generated from SfM can contain considerable distortion due to the orthorectification process, especially near the edges of the reconstruction. Typically, distorted regions are removed to improve the visual consistency of the imagery, as shown in Figure~\ref{fig:orthomosaic}.

\subsection{Dual-Scale Manual Labels}

A ground survey  provided the exact location of 2,929 palms in the study region. Upon observation, we found that many of these palms are not visible in the orthomosaic image, due to severe occlusion or overlap with surrounding trees, see e.g. Figure~\ref{fig:palm_images}. The resulting number of sufficiently visible palms in the orthomosaic was deemed too small for supervised classification. Thus, the following strategy was adopted to generate labeled data.

%%%%
%The centers of palm trees in the UAV images, identified via a ground survey, guided us in cropping 264 by 262 pixel sections around these points to effectively capture the entire palm crown. 
%Yet, as shown in Figure \ref{fig:palm_images}, these sections sometimes miss parts of the palms, especially when they are obscured by other trees or overlap with each other. Such issues compromise the image clarity, prompting us to find a better way to capture the unique structure of palm leaves and differentiate them from other tree crowns.%%%

\paragraph{Fine-scale labeled data.} We used the location of visible palms in the orthomosaic to extract 6000 small image patches of 40 by 40 pixels, capturing as much palm leaf feature diversity as possible. We endeavored to ensure that the extracted patches contain at least 90\% of palm data within them. Similarly, we extracted 6000 40-by-40-pixel patches of non-palm features, such as other types of tree crowns, ground soil, roads, tree trunks, and other land cover types. 

%Consequently, we chose smaller image patches, segmenting the images into 40 by 40 pixel sections from aforementioned images, as this size is adequately detailed to capture the unique characteristics of palm features and to distinguish between palm and non-palm areas (see Figure~\ref{fig:palm_nonpalm_patches}). We selected patches where palms covered at least 90\% of the area to form a palm class and patches devoid of any palm elements to create a contrasting non-palm class. This strategy resulted in approximately 6,000 instances of each class. The non-palm category comprises various elements, including other types of tree crowns, ground, roads, trunks, and other land cover types.

\begin{figure}[tb]
    \centering
    \begin{subfigure}{0.49\linewidth}  % Reduced from 0.8\textwidth
        \centering
        \includegraphics[width=\textwidth]{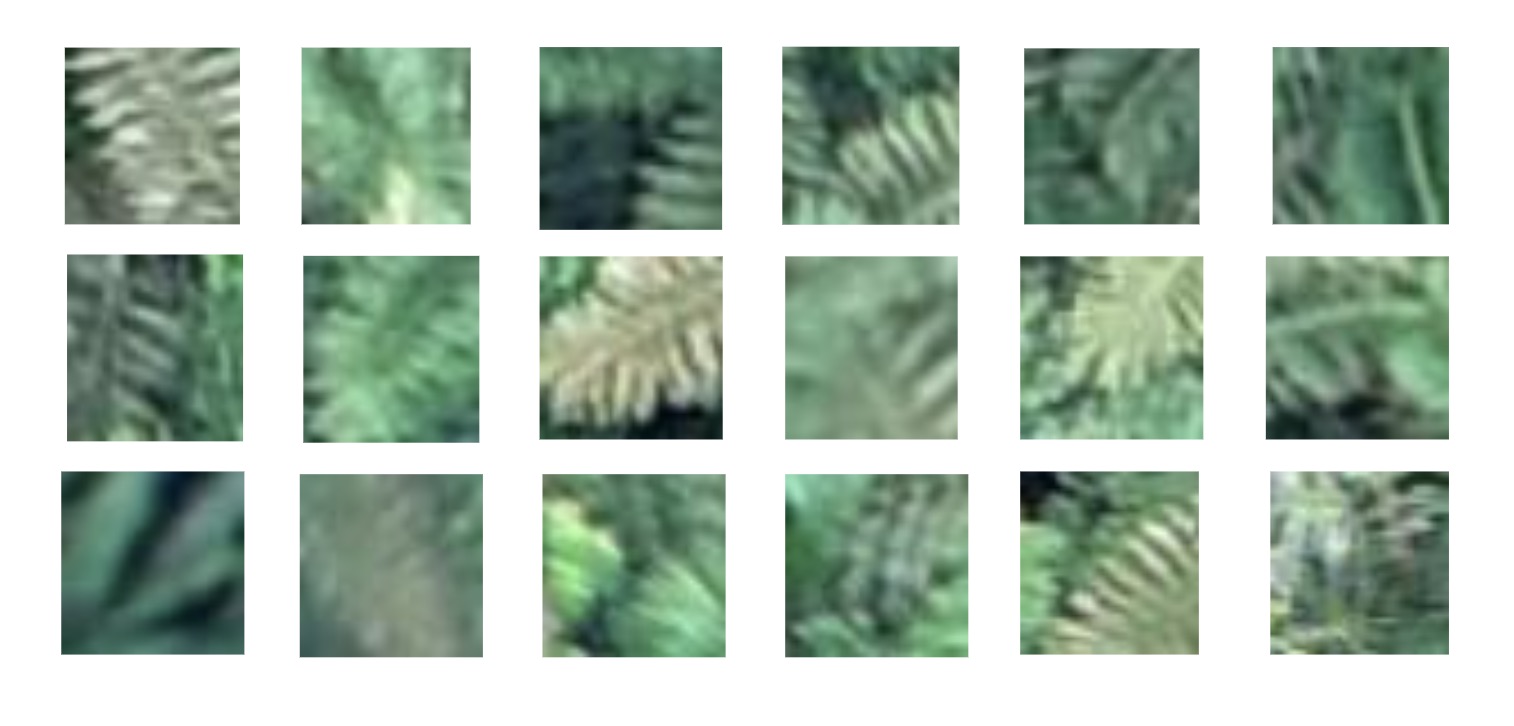}
        \caption{Palms}
    \end{subfigure}
    \begin{subfigure}{0.49\linewidth}  % Reduced from 0.8\textwidth
        \centering
        \includegraphics[width=\textwidth]{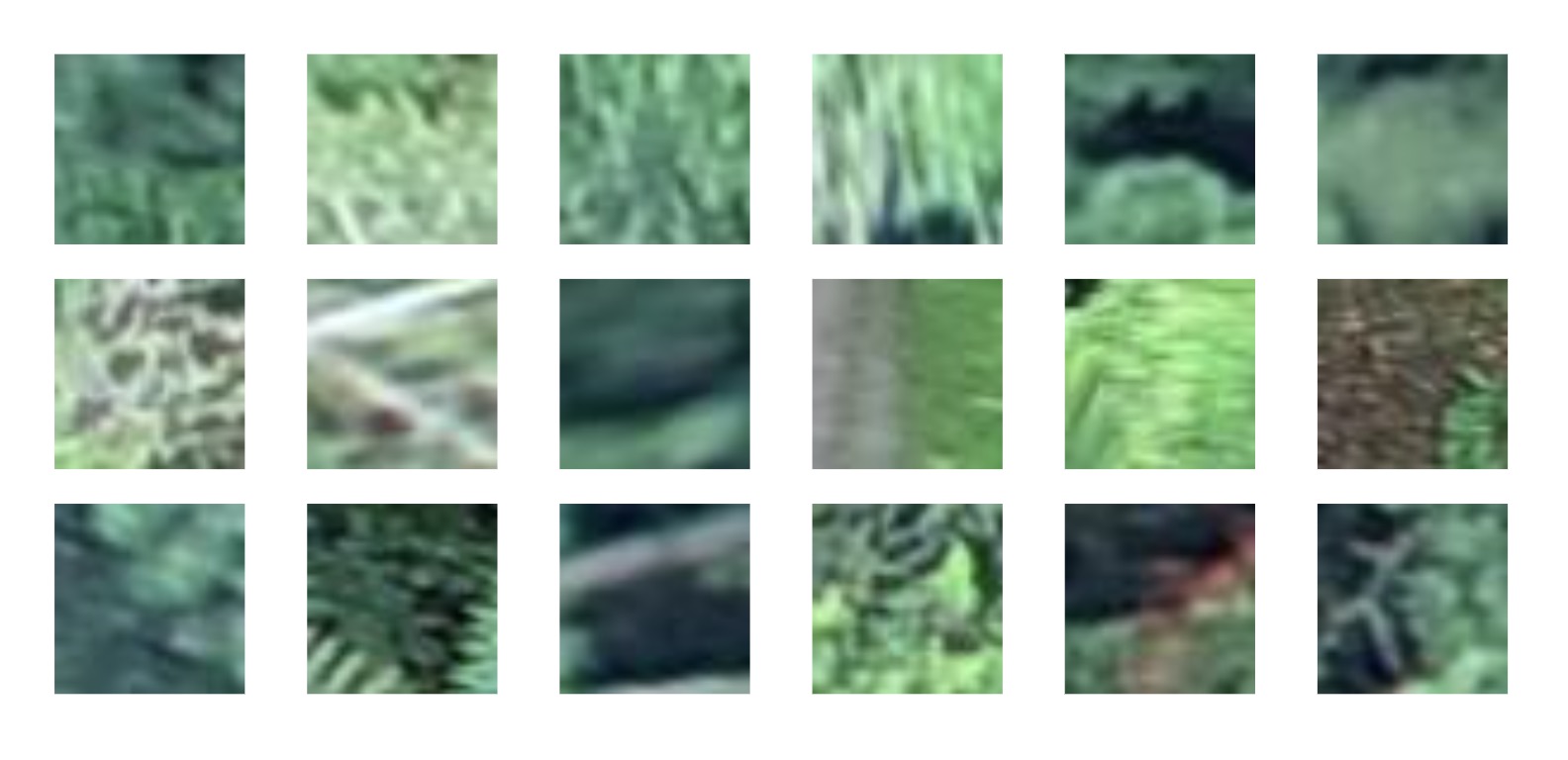}
        \caption{Non-Palms}
    \end{subfigure}
    \caption{Comparison of Small Patches}
    \label{fig:palm_nonpalm_patches}
\end{figure}

Sample images for the resulting palm and non-palm classes are shown in Figure \ref{fig:palm_nonpalm_patches}. The images on the left are patches containing palm features, capturing the distinctive characteristics of a palm tree. The images on the right, on the other hand, showcase patches without palm features, highlighting variations that may help the model to differentiate between palms and non-palms.

\paragraph{Coarse-scale labeled data.}
The second part of our labeling process involved larger 100x100 pixel patches labeled as having high/low probability of containing palms. This scheme not only captures intricate leaf details but also broader tree crown contexts, crucial for distinguishing partially occluded palms from other trees. 

\begin{figure}[tb] % Use figure* for spanning two columns in a two-column document
    \centering
    \begin{subfigure}{0.12\linewidth}
        \includegraphics[width=\textwidth]{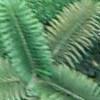}
    \end{subfigure}\hfill%
    \begin{subfigure}{0.12\linewidth}
        \includegraphics[width=\textwidth]{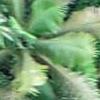}
    \end{subfigure}\hfill%
    \begin{subfigure}{0.12\linewidth}
        \includegraphics[width=\textwidth]{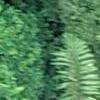}
    \end{subfigure}\hfill%
    \begin{subfigure}{0.12\linewidth}
        \includegraphics[width=\textwidth]{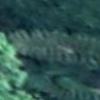}
    \end{subfigure}\hfill%
    \begin{subfigure}{0.12\linewidth}
        \includegraphics[width=\textwidth]{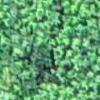}
    \end{subfigure}\hfill%
    \begin{subfigure}{0.12\linewidth}
        \includegraphics[width=\textwidth]{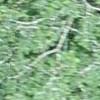}
    \end{subfigure}\hfill%
    \begin{subfigure}{0.12\linewidth}
        \includegraphics[width=\textwidth]{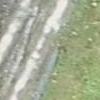}
    \end{subfigure}\hfill%
    \begin{subfigure}{0.12\linewidth}
        \includegraphics[width=\textwidth]{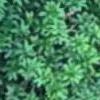}
    \end{subfigure}
    % Subtitles for images
    \small{\hspace{0.09\linewidth}\textbf{(a) Palms}\hspace{0.31\linewidth}\textbf{(b) Non-Palms}}
    \caption{Comparison of Large Patchs}
    \label{fig:large_images}
\end{figure}

A semi-manual approach was employed for this purpose. We used a portion of our fine-scale labeled data to train a network and identify small palm patches throughout the entire orthomosaic image. The locations of these patches reveals the presence of a palm tree, even in areas where the palm is mostly occluded. We then explore 100x100 pixel areas around these patches, labeling them as either palms or non-palms. This approach resulted in about 3367 palm and 4091 non-palm samples, as shown in Figure \ref{fig:large_images}, with the left four images showing palm patches and the right four showing non-palm patches, including roads and other tree crowns. Notably, some palms emerged within the low probability patches and some non-palms within the high probability ones, underscoring the effectiveness of the large patch datasets in tree crown identification.

\section{Methodology}
\label{sec: pipeline}

In this section, we introduce the proposed PalmProbNet that focuses on generating a heatmap to show the probability distribution of palm presence, aiding in biological applications. PalmProbNet encompasses three pivotal stages: feature extraction, classification, and application to the landscape orthomosaic image. The subsequent sections provide a detailed introduction of these stages and Figure~\ref{fig:workflow} shows the workflow of the proposed PalmProbNet.

\subsection{Feature Extraction}

Transfer learning is a potent technique that leverages pre-trained models, reducing the need for extensive data collection and computational resources \cite{weiss2016survey}. Customization of these models is often achieved by adding specialized layers tailored to the specific task. Multi-Layer Perceptrons (MLPs), feedforward neural networks with multiple layers, are commonly used for this purpose due to their capability to learn intricate patterns \cite{rumelhart1985learning}. Residual Networks (ResNet) address challenges in training deep neural networks by introducing shortcut or "skip" connections, making them suitable for complex tasks like palm tree detection~\cite{he2016deep}. In our study, given the limited labeled data, we utilized the pre-trained ResNet-18 model, a variant with 18 layers. Originally trained on the ImageNet dataset~\cite{deng2009imagenet}, ResNet-18 offers a robust feature hierarchy. We fine-tuned its final layers to our dataset, ensuring a robust foundation for the PalmProbNet. Additionally, we incorporated data augmentation techniques to enhance the diversity of our training data, addressing issues such as variable illumination and occlusions~\cite{liu2021automatic}.

\subsection{Classification}

\begin{figure}[tb] 
    \centering
    \includegraphics[width=0.9\linewidth]{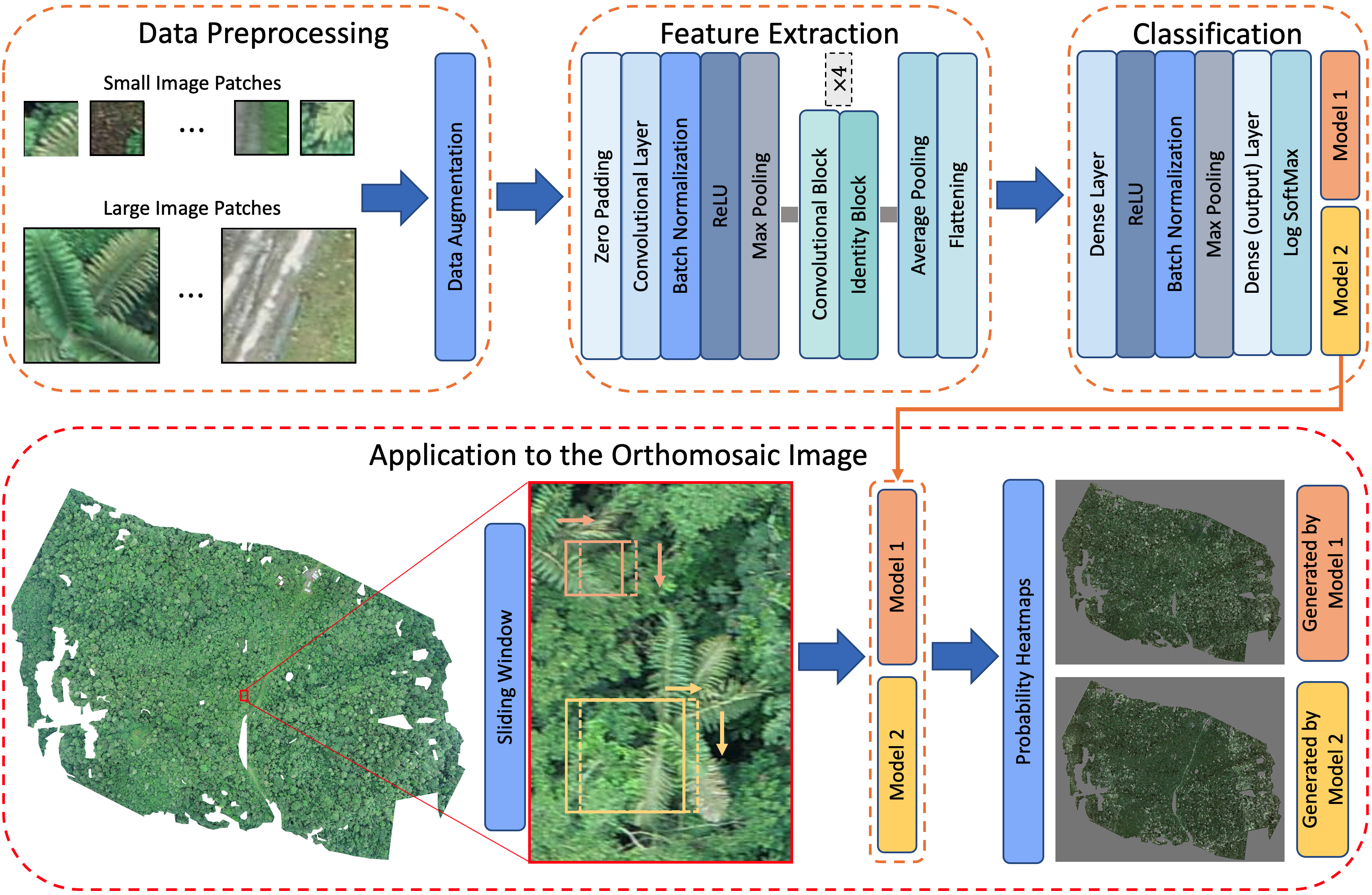}
    \caption{Workflow of The Proposed PalmProbNet}
    \label{fig:workflow}
\end{figure}

To classify the palm images based on the extracted features from ResNet layers, PalmProbNet employs an MLP that can capture complex patterns in data by introducing non-linearity~\cite{rumelhart1985learning}. Our MLP architecture, shown in Figure \ref{fig:workflow}, incorporates a batch normalization layer, enhancing the model's ability to learn from ResNet's feature vectors and subsequently map them to the appropriate class labels. The overall workflow is illustrated in Figure \ref{fig:workflow}.

\subsection{Application to the Orthomosaic Image}\label{sec: application}

Applying the model trained by small/large patches to large landscape orthomosaic images in PalmProbNet involves several steps. Depending on the training set, the model uses either \(40 \times 40\) or \(100 \times 100\) pixel patches. The process begins by loading the image, preprocessing, and capturing patches using a sliding window with a stride of 10. Patches with 25\% or more missing pixels are immediately categorized as `non-palm'. The ResNet layers then extracts features from each patch, which are predicted by the trained MLP layers. Probabilities from these predictions are aggregated into an accumulator array, with overlapping regions being averaged to ensure accurate probability estimates. Following the predictions, the averaged probabilities are resized to generate a heatmap, which is then overlaid onto the orthomosaic image to emphasize the areas containing palms. The final results, including the highlighted image indicating the palm regions and the heatmap of the landscape orthomosaic, are saved for further analysis, capturing all computed probabilities.

%% Experiment section

\section{Experiment}\label{sec: numerics}

\subsection{Data Division and Augmentation}

To ensure a fair evaluation of PalmProbNet, we divided our dataset (for both small and large patches datasets) into two parts: 80\% for training and 20\% for testing. The training set was further subjected to 5-fold cross-validation, providing a more reliable assessment of the model's performance.

Data augmentation enhances model robustness and performance, especially in image-based tasks with limited labeled datasets. For the training set, we implemented several preprocessing and augmentation techniques. The images were resized to \(224 \times 224\) pixels for compatibility with the pre-trained ResNet layers. Augmentations included random horizontal and vertical flips, which mirror the image in different orientations, and color jittering to simulate various lighting conditions. Lastly, normalization was applied to scale pixel values to a standardized range, facilitating faster convergence during training. For the test set, we avoided augmentations that might alter the intrinsic characteristics of the images, preserving consistency with the model's expectations.

\subsection{Model Training and Evaluation}

In the training of PalmProbNet, we fine-tuned the MLP layer parameters for task-specific adaptations, while keeping most ResNet parameters frozen to guarantee robust feature extraction. The final layer of ResNet was specifically fine-tuned to better align the network's output with the unique characteristics of our dataset. During each fold of the cross-validation, the training data was augmented using the described techniques, while the validation data remained unchanged. We submitted SLURM jobs on Wake Forest University's cluster to accelerate the computation. The training employed the Negative Log-Likelihood loss function to evaluate prediction accuracy.

The core training phase consisted of 500 epochs for small patches and 200 epoches for large patches, employing a batch size of 64. We monitored the validation loss to select the best model, characterized by the lowest loss across epochs and folds. This model was then retrained on the entire training dataset with the same augmentation techniques and evaluated on a non-augmented test set. This systematic training approach, integrating cross-validation and validation loss monitoring, provided a robust evaluation framework, ensuring optimal model performance and reliability.

We utilized several metrics to evaluate the performance of PalmProbNet. Accuracy, given by $\text{Acc} = \frac{\text{Number of Correct Predictions}}{\text{Total Predictions}}$, measures the proportion of correct classifications. Average Accuracy (AA) is the mean of individual class accuracies, offering a balanced view, especially for imbalanced datasets. Cohen's $\kappa$ coefficient gauges prediction reliability by assessing agreement between predicted and actual classes beyond chance. For binary classification, Precision, defined as $\text{Precision} = \frac{\text{True Positives}}{\text{True Positives + False Positives}},$ evaluates the model's exactness, while Recall, given by $\text{Recall} = \frac{\text{True Positives}}{\text{True Positives + False Negatives}},$ measures its completeness. The ROC AUC represents the model's discriminative power between classes.

\begin{figure}[h]
    \centering
    \begin{subfigure}{0.45\linewidth}
        \includegraphics[width=\textwidth]{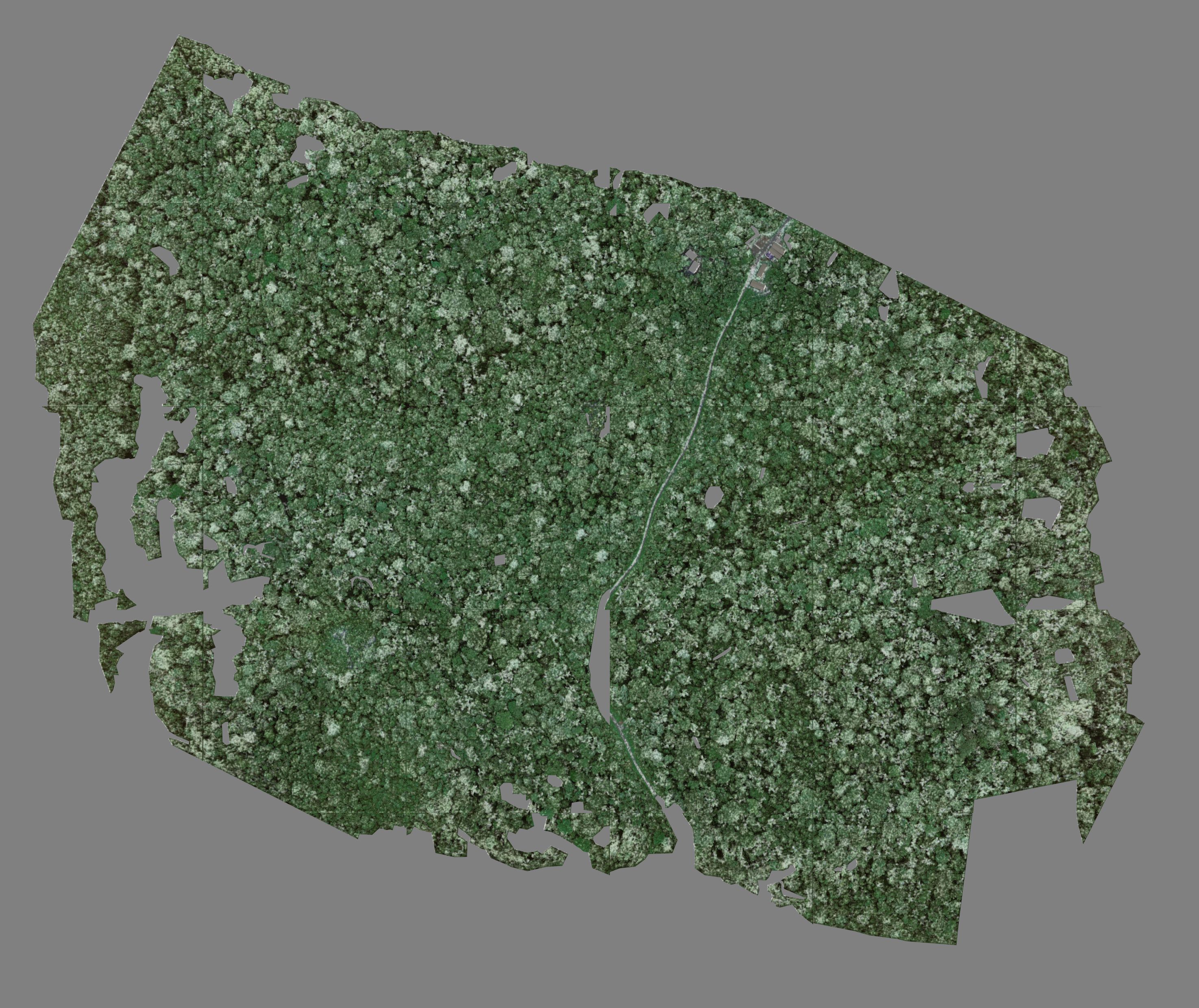}
        \caption{With \texttt{nnodes} = 64, Small}
        \label{subfig:prediction_map_64}
    \end{subfigure}
    \begin{subfigure}{0.45\linewidth}
        \includegraphics[width=\textwidth]{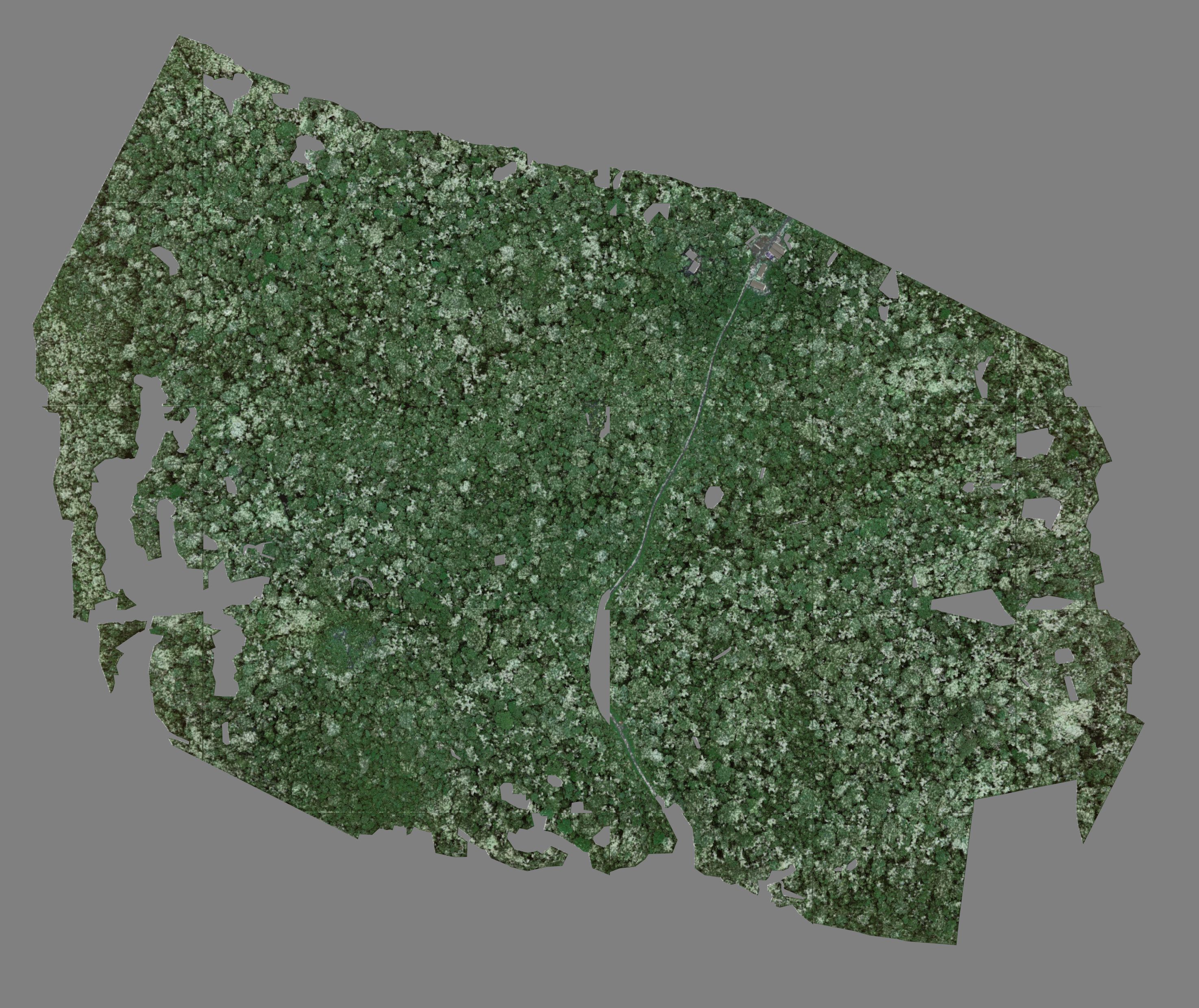}
        \caption{With \texttt{nnodes} = 128, Small}
        \label{subfig:prediction_map_128}
    \end{subfigure}
    
    \begin{subfigure}{0.45\linewidth}
        \includegraphics[width=\textwidth]{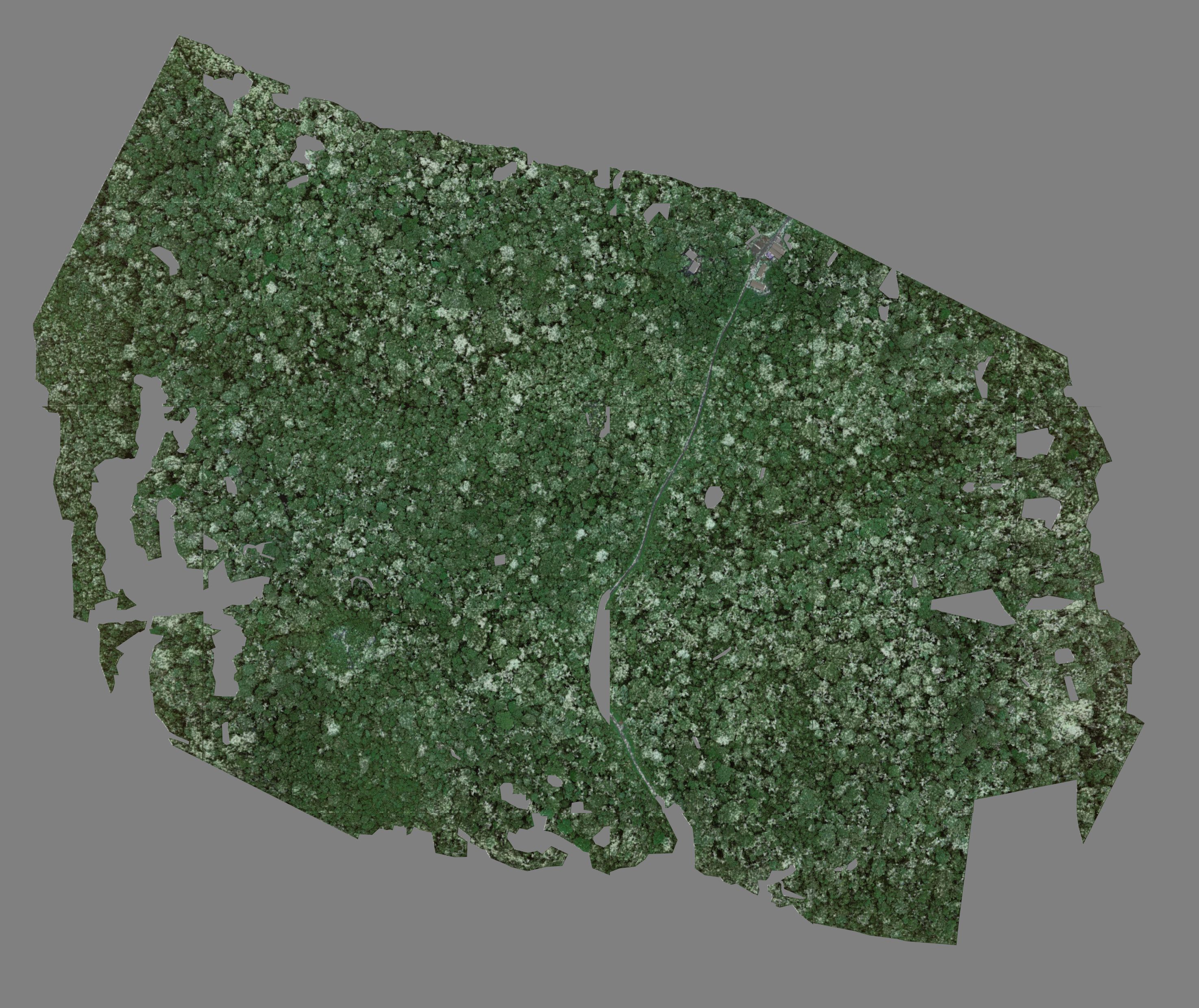}
        \caption{With \texttt{nnodes} = 256, Small}
        \label{subfig:prediction_map_256}
    \end{subfigure}
    \begin{subfigure}{0.45\linewidth}
        \includegraphics[width=\textwidth]{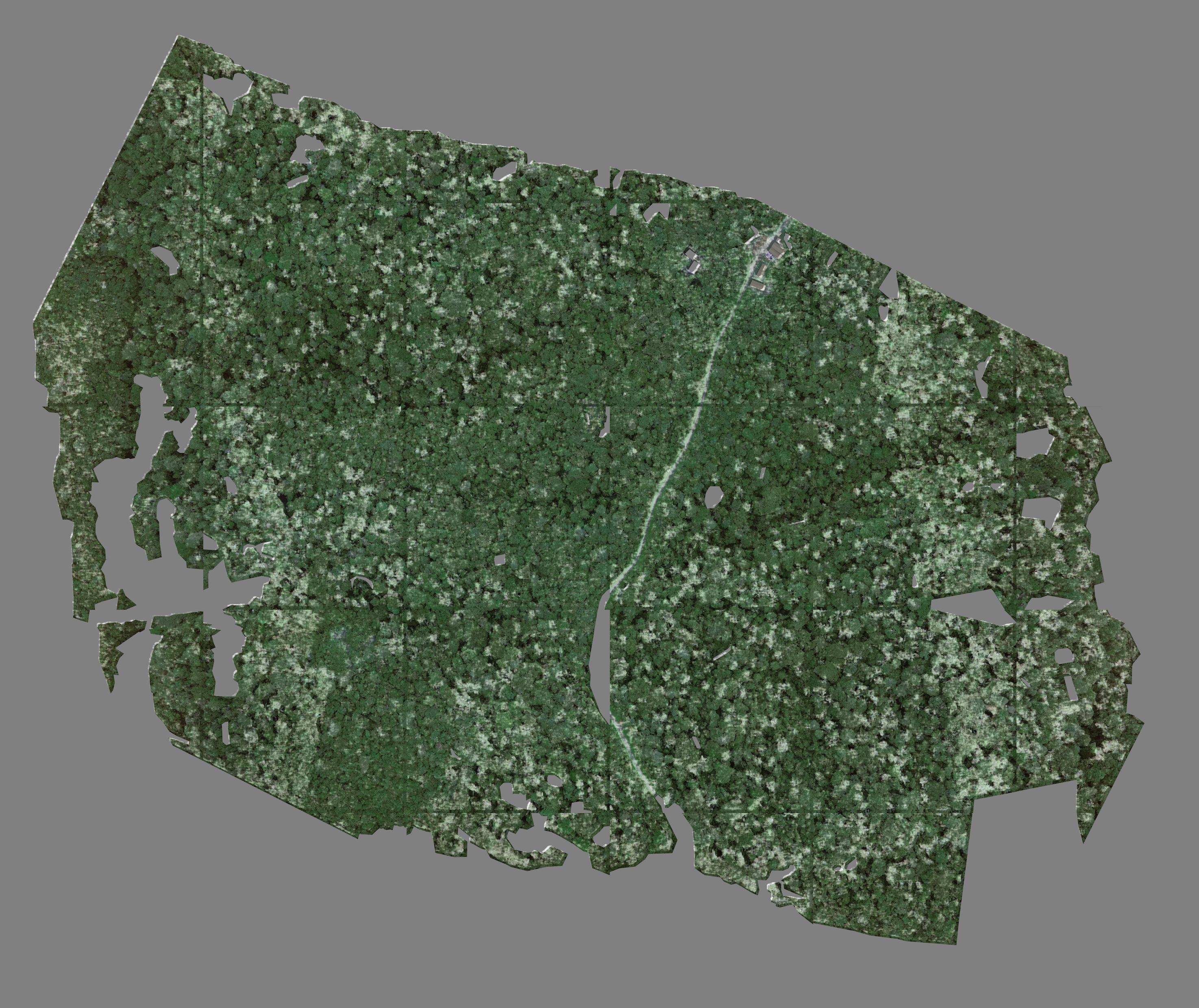}
        \caption{With \texttt{nnodes} = 64, Large}
        \label{subfig:large_prediction_map_64}
    \end{subfigure}
    
    \begin{subfigure}{0.45\linewidth}
        \includegraphics[width=\textwidth]{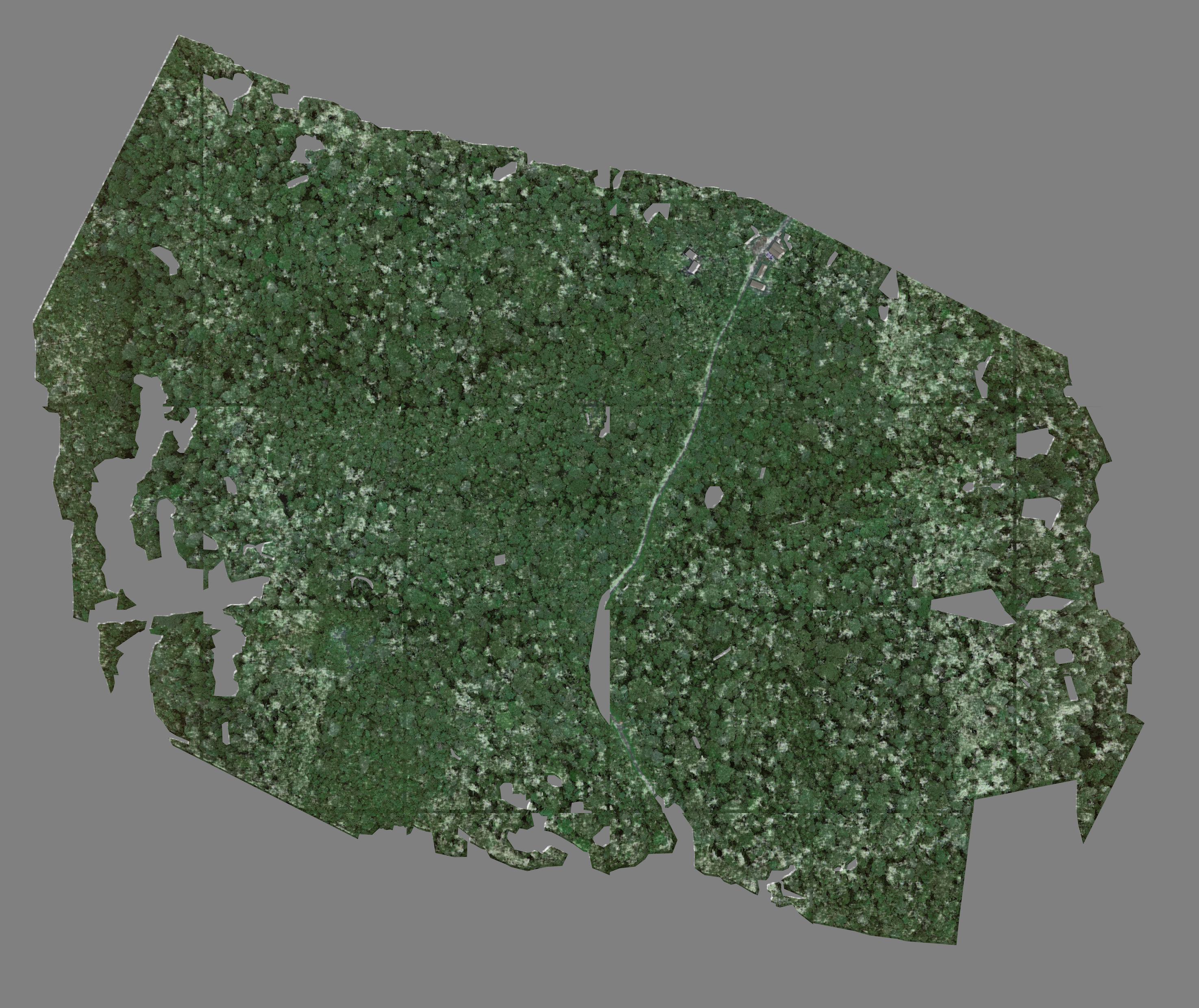}
        \caption{With \texttt{nnodes} = 128, Large}
        \label{subfig:large_prediction_map_128}
    \end{subfigure}
    \begin{subfigure}{0.45\linewidth}
        \includegraphics[width=\textwidth]{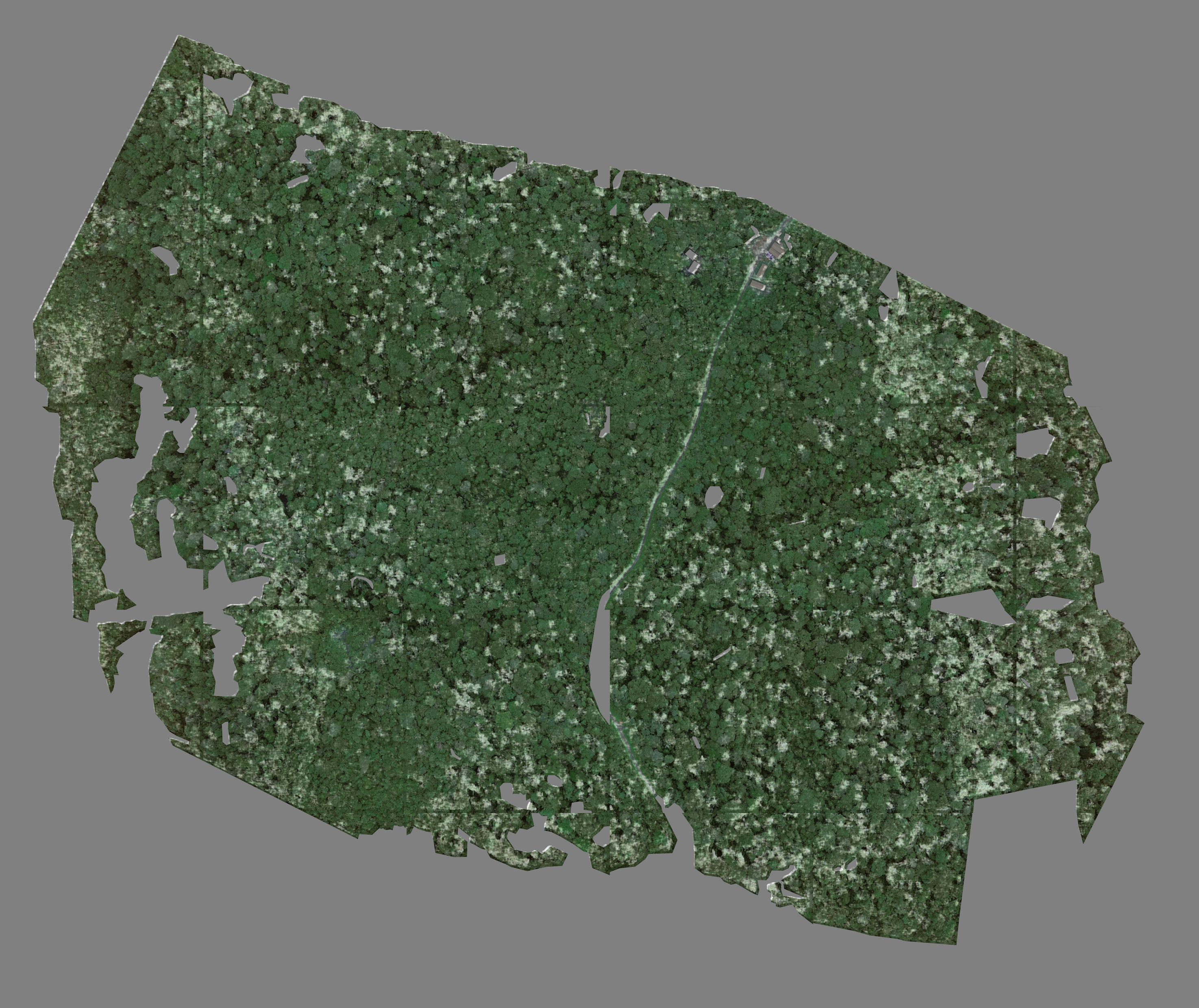}
        \caption{With \texttt{nnodes} = 256, Large}
        \label{subfig:large_prediction_map_256}
    \end{subfigure}
    \caption{Probability Heatmaps Produced by PalmProbNet Trained on Small \& Large Patches}
    \label{fig:prediction_maps}
\end{figure}

\subsection{Numerical Results and Discussions}

In evaluating the numerical results, the trained models tailored for both large and small patch sizes were tested on the testing patches and deployed on the landscape orthomosaic image. A stride of 10 was employed during this process, which allowed for a comprehensive and efficient scan of the entire image.

Our comparative analysis of patch classification, encompassing both small and large patches, involved three configurations of hidden layer nodes (\texttt{nnodes}): 64, 128, and 256. The performances, shown in Table \ref{tab:combined_patches_results}, reveal a uniform level of effectiveness across the configurations for both patch sizes. For small patches, the 256-node configuration showed a slight edge, although the overall stable metrics across all configurations indicate that additional fine-tuning may be necessary for significant performance improvements. In the realm of large patches, the 256-node setup similarly displayed a marginal but noticeable enhancement in performance measures, such as accuracy, average accuracy, and Cohen's $\kappa$ coefficient. These parallel findings across different patch sizes underscore the robustness of the classification approach and suggest that the 256-node configuration provides a subtle yet consistent advantage.

\begin{table}[tb]
\centering
\caption{Classification Results for Different Configurations}
\label{tab:combined_patches_results}
\resizebox{0.7\linewidth}{!}{%
\begin{tabular}{c|c|ccccccc|}
\hline
\multicolumn{1}{c|}{Patch Size} & \texttt{nnodes} & Acc & ROC AUC & AA & $\kappa$ & Precision & Recall \\ \hline
\multirow{3}{*}{Small} & 64  & 0.9101 & 0.9642 & 0.9105 & 0.8203 & 0.9126 & 0.9101 \\
                       & 128 & 0.9041 & 0.9653 & 0.9047 & 0.8084 & 0.9099 & 0.9041 \\
                       & 256 & \textbf{0.9110} & \textbf{0.9674} & \textbf{0.9114} & \textbf{0.8221} & \textbf{0.9132} & \textbf{0.9110} \\ \hline
\multirow{3}{*}{Large} & 64  & 0.9698 & \textbf{0.9963} & 0.9694 & 0.9391 & 0.9698 & 0.9698 \\
                       & 128 & 0.9712 & 0.9949 & 0.9705 & 0.9418 & 0.9712 & 0.9712 \\
                       & 256 & \textbf{0.9732} & 0.9959 & \textbf{0.9735} & \textbf{0.9459} & \textbf{0.9733} & \textbf{0.9733} \\ \hline
\end{tabular}
}
\end{table}

Figure~\ref{fig:prediction_maps}(a-c) presents the probability heatmaps generated from the landscape orthomosaic image using the PalmProbNet trained on small patches. These heatmaps visually encode the model's assessed probability of palm presence, with brighter areas corresponding to higher probabilities of palm features. The intensity within each \(10 \times 10\) segment of the heatmap is indicative of the model's confidence in identifying palm elements within that specific area.

Considering three different hidden layer configurations (\texttt{nnodes} = 64, 128, 256), the results, while broadly similar, exhibit subtle differences upon closer inspection. Each configuration efficiently detects a majority of the palm trees, though there are occasions where non-palm tree crowns are erroneously classified as palms. Such instances emphasize the necessity for more refined model adjustments to reduce false positives. A visual evaluation suggests that configurations with \texttt{nnodes} = 64 and \texttt{nnodes} = 256 marginally surpass the performance of \texttt{nnodes} = 128. The consistent detection patterns across diverse node configurations underline the reliability of the feature extraction and classification phases, simultaneously indicating areas where further tuning could elevate model accuracy.

Figure~\ref{fig:prediction_maps}(d-f) displays the probability heatmaps derived from the landscape orthomosaic image, utilizing PalmProbNet trained on large patches. Each map is the result of a different hidden layer node configuration (\texttt{nnodes} = 64, 128, 256), with the intensity of the highlights indicating the likelihood of palm presence. The three configurations yield very similar results, making it challenging to determine a clear superior performer. However, a common issue across all configurations is the misclassification of certain areas. Notably, patches adjacent to blank areas---those that have been either deleted due to poor image quality or are missing---are prone to be inaccurately labeled as containing palms. Addressing these misclassifications may require incorporating training samples that include these edge cases or applying post-processing steps to exclude them, thereby refining the model's ability to discern true palm features from artifacts introduced by image processing.

\section{Conclusion and Future Work}\label{sec: conclusions}

Identification of ecologically and economically important species within tropical forests canopies expands the frontiers of research and conservation. The paper details a thorough investigation into identifying palm trees in Neotropical rain forest using PalmProbNet on UAV-captured images. In particular, it tackles challenges commonly encountered in detection tropical palm canopies like noise and uneven lighting, providing precise palm localization and landscape spatial distributions. PalmProbNet employed two models with the same architecture but trained on different-sized image patches. The models, particularly with a 256-node configuration, yielding robust results: for small patches, an accuracy of 0.9110, ROC AUC of 0.9674, and a Cohen's $\kappa$ coefficient of 0.8221; for large patches, an accuracy of 0.9732, ROC AUC of 0.9959, and a Cohen's $\kappa$ coefficient of 0.9459. Probability heatmaps were introduced as a new way to illustrate palm distributions, useful for both environmental and economic research and computational tasks. The integration of UAVs and deep learning has been validated for palm detection in dense forests through high classification accuracy across various model configurations.

Future enhancements to PalmProbNet will focus on refining the heatmap by including training samples with edge cases and employing post-processing to filter out patches near blank areas. We plan to include a fairer comparison of PalmProbNet with other classifiers, and explore segmentation networks using the produced heatmap. We also intend to involve superpixel-based methods for the fast and precise localization of individual palms in orthomosaic images, enhancing our understanding of their distribution~\cite{cui2023superpixel}. Additionally, we aim to augment our approach by integrating UAV imagery with satellite data from platforms like Planet or WorldView, capitalizing on the complementary strengths of local detail and broad-scale context to improve detection accuracy~\cite{camalan2022detecting}.

\printbibliography 

\end{document}